\documentclass[10pt]{article} 
\usepackage[accepted]{tmlr}

\usepackage{hyperref}
\usepackage{url}

\usepackage{graphicx}
\usepackage{booktabs}
\usepackage{xcolor}
\usepackage{multirow}
\usepackage{amsmath,amssymb,amsfonts}
\usepackage{mathrsfs}
\usepackage[title]{appendix}
\usepackage{textcomp}
\usepackage{manyfoot}
\usepackage{algorithm}
\usepackage{algpseudocode}
\usepackage{listings}
\usepackage{pifont}
\usepackage{soul}
\usepackage{enumitem}
\usepackage{tabularx}
\usepackage{lipsum}
\usepackage{mathbbol}
\usepackage{tabularray}
\usepackage[utf8]{inputenc}
\usepackage{color, colortbl}
\usepackage{wrapfig}
\usepackage{placeins}
\usepackage{caption}

\newcommand{\revise}[1]{\textcolor{black}{#1}}
\newcommand{\reviseb}[1]{\textcolor{black}{#1}}
\newcommand{\reviseC}[1]{\textcolor{black}{#1}}
\usepackage{xspace}

\usepackage[capitalize]{cleveref}
\crefname{section}{Section}{Sections}
\crefname{table}{Table}{Tables}
\crefname{figure}{Figure}{Figures}

\newcommand{\cmark}{\textcolor{green}{\ding{51}}}
\newcommand{\xmark}{\textcolor{red}{\ding{55}}}
\newcommand{\model}{\textsf{{REVEAL}}\xspace}
\newcommand{\dataset}{\textsf{{LAVAT}}\xspace}

\title{REVEAL: Reconstructing Video Amodal Content via Language and Flow Guidance}

\author{\name Minh Tran \email minht@uark.edu \\
      \addr Department of Electrical Engineering and Computer Science\\
      University of Arkansas
      \AND
      \name Winston Bounsavy \email bounsavy@uark.edu \\
      \addr Department of Electrical Engineering and Computer Science\\
      University of Arkansas
      \AND
      \name Ngan Le \email thile@uark.edu\\
      \addr Department of Electrical Engineering and Computer Science\\
      University of Arkansas}

\def\month{11}  
\def\year{2026} 
\def\openreview{\url{https://openreview.net/forum?id=pbTFy1bNyZ}} 

\begin{document}
\maketitle
\begin{center}
    \includegraphics[width=.85\linewidth]{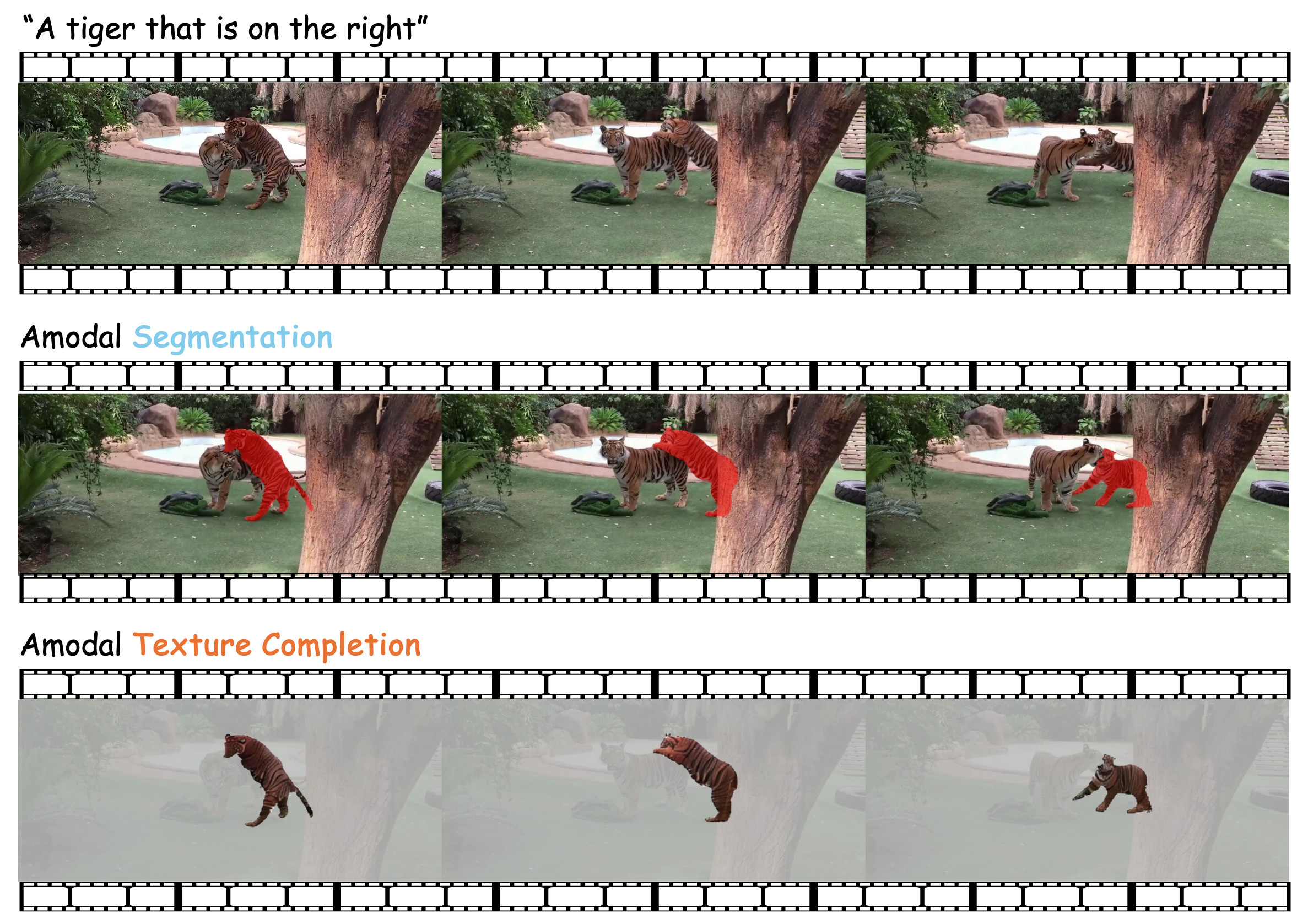}
    \captionof{figure}{We tackle the problem of open-vocabulary video amodal segmentation and content completion: given an input video and a text prompt, \model generates amodal (visible + invisible) masks and RGB content via a two-stage pipeline guided by optical flow and language priors.}
    \label{fig:teaser}
\end{center}

\begin{abstract}
Amodal perception enables humans to perceive entire objects even when parts are occluded, a remarkable cognitive skill that artificial intelligence struggles to replicate.
Recent diffusion-based methods have extended amodal completion from images to videos, yet they lack auxiliary priors to guide the reconstruction of heavily occluded content and to maintain temporal consistency over long sequences.
We present \textbf{\model} (\textbf{RE}constructing \textbf{V}id\textbf{E}o \textbf{A}modal content via \textbf{L}anguage and flow guidance), a unified diffusion-based framework that integrates complementary motion and semantic priors for video amodal completion. A user-provided text query serves a dual role: it drives open-vocabulary video segmentation to obtain visible masks, and it provides semantic guidance for texture reconstruction. For amodal segmentation, we introduce optical flow guidance: by warping visible masks from previous frames to the current frame, the flow-warped mask propagates visible content into occluded regions and approximates the object's current shape, providing a strong shape prior even under simultaneous deformation and occlusion. For texture completion, text guidance constrains the reconstruction of occluded appearance while providing a stable semantic reference that helps maintain visual coherence throughout the reconstruction.
We also contribute \textbf{\dataset}, a new benchmark featuring long video sequences paired with text descriptions, enabling evaluation of text-guided video amodal completion under heavy occlusion. Extensive experiments demonstrate that \model achieves state-of-the-art performance on existing benchmarks while maintaining high-quality temporal consistency over extended sequences. The project page is available at: \href{https://uark-aicv.github.io/REVEAL/}{https://uark-aicv.github.io/REVEAL/}.
\end{abstract}

\section{Introduction}
\label{sec:intro}

Humans possess an extraordinary ability to perceive objects as complete entities, even when parts are obscured. In everyday life, objects frequently occlude one another, yet we effortlessly infer and reconstruct their hidden regions, a capability known as amodal perception or amodal completion~\citep{kellman1991theory, van1999investigating}. The human visual system relies on cues such as shape continuity and symmetry~\citep{van1999investigating}, as well as extensive prior knowledge of the physical world~\citep{yun2018temporal}. Replicating this cognitive process in artificial intelligence (AI) remains highly challenging, despite significant advances in computer vision.

Enabling amodal completion in AI would benefit numerous applications, including robotics~\citep{qin2020s4g,back2022unseen}, autonomous driving~\citep{qi2019amodal}, and augmented reality~\citep{ozguroglu2024pix2gestalt}. Inspired by human perception, early research primarily focused on amodal segmentation~\citep{li2016amodal,follmann2018mvtec,follmann2019learning,tran2022aisformer,yao2022self,gao2023coarse}, aiming to recover the full object mask. More recently, denoising diffusion models~\citep{ho2020denoising,rombach2022high} have advanced amodal content completion for images~\citep{ozguroglu2024pix2gestalt,xu2024amodal,zhan2024amodal} and subsequently for videos~\citep{chen2025using, lu2025taco}.

Despite this progress, recent video amodal completion methods often lack auxiliary priors to guide the reconstruction of occluded regions. This limitation becomes particularly critical under heavy occlusion, where inferring hidden content is inherently ambiguous or even ill-posed~\citep{fan2023rethinking, gao2023coarse, yao2022self}, as well as in long video sequences, where such priors are essential for maintaining temporal consistency throughout the reconstruction~\citep{yao2022self}.

For video amodal segmentation, motion information is a particularly valuable source of prior knowledge. Optical flow encodes how visible regions move between frames, so warping a previous frame's mask with the estimated flow can propagate visible content into occluded areas of the current frame, providing a direct shape prior. This is especially beneficial when an object simultaneously deforms and becomes occluded: in such cases, the object's shape in earlier frames no longer matches its current shape, making appearance-based cues unreliable. Optical flow addresses this by capturing the underlying motion field, allowing the model to approximate the object's current shape even as it changes.

For amodal texture completion, the appearance of occluded regions must be reconstructed consistently across the entire video. Semantic priors, such as language descriptions, provide high-level guidance about the object's category and attributes, constraining the reconstruction of hidden content. Moreover, since the text description remains constant across all frames, it provides a stable semantic reference that helps maintain visual coherence throughout the reconstruction.

In this work, we propose \model (\textbf{RE}constructing \textbf{V}id\textbf{E}o \textbf{A}modal content via \textbf{L}anguage and flow guidance). 
Our framework integrates complementary motion and semantic priors within a diffusion-based architecture. Instead of requiring a visible mask as input, \model allows the user to provide a text query specifying the object of interest. The text query plays a dual role in our pipeline: it drives open-vocabulary video segmentation (e.g., SAM3~\citep{carion2025sam}) to obtain visible masks across frames, and it serves as a semantic prior in the texture completion stage, guiding the reconstruction of occluded content with object-class knowledge. This end-to-end integration of text makes the framework open vocabulary (\cref{fig:teaser}). 

To incorporate motion priors, we introduce optical flow guidance. Specifically, we warp the visible mask from a previous frame to the current frame using the estimated optical flow, producing a flow-warped mask that approximates the object's full extent in the current frame. Because the warped mask can recover regions that are currently occluded but were previously visible, it provides a strong shape prior for amodal mask prediction. Furthermore, when occlusion and non-rigid deformation occur simultaneously, the flow-warped mask captures the object's shape transformation, offering a reliable cue where static shape priors from earlier frames would fail.
To incorporate semantic priors, we leverage textual description guidance. Text features encode semantic attributes and canonical shape information of the target category, constraining the reconstruction of occluded content. By combining motion consistency with semantic understanding, our framework provides stronger guidance than previous methods, improving performance and consistency under severe occlusion and in long videos.

In addition, we present {\dataset}, a new benchmark for text-guided video amodal completion. {\dataset} contains long video sequences paired with text descriptions, enabling evaluation of both temporal consistency and text-video alignment. It addresses the limitations of prior benchmarks that lack text annotations and focus exclusively on short clips.

In summary, our contributions are:
(1) We introduce \model, a unified framework that integrates optical flow as a motion prior and language as a semantic prior for video amodal completion, addressing the lack of auxiliary guidance under heavy occlusion.
(2) We introduce \dataset, a new benchmark for text-guided video amodal completion, featuring long video sequences with text descriptions for comprehensive evaluation. Extensive experiments demonstrate that \model achieves state-of-the-art performance on video amodal completion benchmarks.

\section{Related Work}
\label{sec:related_work}

\subsection{Amodal Completion and Segmentation} \label{sec:related:completion_segmentation}
Amodal visual understanding addresses scenarios where objects are partially hidden by foreground elements. Amodal segmentation recovers the full object mask including occluded regions~\citep{zhu2017semantic,ke2021deep,qi2019amodal,reddy2022walt,ling2020variational}, while amodal detection predicts the complete object extent~\citep{kar2015amodal,hsieh2023tao}. Progress in mask-based modeling and the availability of closed-world~\citep{zhan2020self,ling2020variational,ke2021deep,qi2019amodal,kar2015amodal,zhu2017semantic} and synthetic datasets~\citep{ehsani2018segan} have fueled strong methods such as PCNet~\citep{zhan2020self} and AISFormer~\citep{tran2022aisformer}.
More recently, amodal completion has emerged, aiming to reconstruct the entire object shape~\citep{ehsani2018segan,zhan2020self,ozguroglu2024pix2gestalt}. pix2gestalt~\citep{ozguroglu2024pix2gestalt} fine-tunes a large diffusion model on synthetic data, while MCAmodal~\citep{xu2024amodal} and OpenAmodal~\citep{ao2025open} reframe the task as inpainting and leverage pretrained models directly.
Building on image-level advances~\citep{ozguroglu2024pix2gestalt,xu2024amodal,zhan2024amodal}, recent efforts extend amodal completion to videos~\citep{chen2025using,lu2025taco}. \textit{However, existing video methods lack sufficient auxiliary cues for recovering severely occluded regions and maintaining temporal consistency over longer sequences. Our \model addresses these limitations by incorporating motion and semantic priors.}

\subsection{Video Diffusion Models}
\label{related:diffusion}
Denoising Diffusion Probabilistic Models (DDPMs)~\citep{ho2020denoising} have emerged as powerful generative methods, applied to image generation~\citep{ho2022classifier,song2020denoising,dhariwal2021diffusion}, video generation~\citep{blattmann2023stable,ho2022imagen,gu2023reuse}, motion generation~\citep{guo2023animatediff,zhang2024avid}, and 3D generation~\citep{liu2023zero,deitke2023objaverse,wu2023sin3dm}. A key milestone was~\citep{dhariwal2021diffusion}, showing diffusion models can surpass GANs~\citep{goodfellow2014generative} in image synthesis, followed by Stable Diffusion~\citep{rombach2022high} which enables text-guided generation and editing~\citep{brooks2023instructpix2pix,gal2022image,ruiz2023dreambooth}.
For video synthesis, VDM~\citep{ho2022video} extends text-to-image diffusion to joint image-video training, Imagen Video~\citep{ho2022imagen} employs cascaded diffusion with super-resolution, and Make-A-Video~\citep{singer2022make} leverages unsupervised video data. Gen-1~\citep{esser2023structure} introduces structure-guided video editing, while Tune-A-Video~\citep{wu2023tune} enables one-shot generation. Recent models such as CogVideoX~\citep{yang2024cogvideox} and WAN-Video~\citep{wan2025wan} support diverse tasks including text-to-video, image-to-video, and video editing. \revise{A complementary line of work conditions these backbones on motion or addresses long-horizon generation: MOFA-Video~\citep{niu2024mofa} warps multi-scale first-frame features under a dense flow field through a ControlNet-style adapter for controllable image-to-video animation, and Self-Forcing~\citep{huang2026self} mitigates exposure bias in autoregressive video diffusion via post-training distillation on a causal-DiT backbone.} \textit{Our approach builds upon video diffusion models to leverage shape and texture priors learned from large-scale pretraining.}

\subsection{\reviseC{Layered Video Decomposition}}
\label{sec:related:video_decomposition}
\reviseC{\noindent
Generative layered-video methods such as Generative Omnimatte~\citep{lee2025generative}, OmnimatteZero~\citep{samuel2025omnimattezero}, or EasyOmnimatte~\citep{hu2026easyomnimatte} reconstruct occluded content while decomposing a video into editable foreground and background layers. While generative layered-video decomposition approaches can reconstruct occluded object content, their primary goal is to decompose a video into editable foreground and background layers, where a foreground layer may also include associated effects such as shadows or reflections. In contrast, our formulation explicitly targets object-centric amodal completion, aiming to recover the complete spatial extent and appearance of a specific object under occlusion. This distinction provides several capabilities that are important for amodal perception. Our method directly produces an amodal object mask and completed object appearance, without requiring the occluder to be represented as a separate layer. It therefore naturally handles cluttered scenes and severe or complete occlusions, while providing a representation that can be directly used for tasks such as amodal segmentation, tracking, and 3D/4D reconstruction.}

\section{Methodology}

\textbf{Problem Setting. } 
Let $\mathbf{I} \in \mathbb{R}^{N \times 3 \times H \times W}$ denote a video sequence and let $\mathbf{p}$ be an initial text prompt (e.g., ``a golden retriever'') specifying the object of interest in the first frame. We first employ an open-vocabulary video segmentation model (e.g., SAM3~\citep{carion2025sam}) to obtain the visible masks $\mathbf{V} \in \mathbb{R}^{N \times H \times W}$ of the object across the entire sequence.  
Our framework consists of two stages. In the first stage, we predict amodal masks $\mathbf{A} \in \mathbb{R}^{N \times H \times W}$ that capture the complete spatial extent of the object, including occluded regions. In the second stage, conditioned on $\mathbf{A}$, we reconstruct the RGB content (texture) of the occluded regions, yielding $\mathbf{C} \in \mathbb{R}^{N \times 3 \times H \times W}$. The objective is for the visible portions of $\mathbf{C}$ to be faithful to the corresponding regions in the input video $\mathbf{I}$, while the completed (occluded) regions should blend naturally with the scene, preserving contextual coherence and avoiding physically implausible object structures.

We adopt a two-stage design because it provides greater stability and consistency. A single-stage approach often struggles to infer complete object shapes, making it even more challenging to generate coherent texture for the occluded regions~\citep{chen2025using}.

\begin{figure*}[!t]
    \centering \includegraphics[width=1.\textwidth]{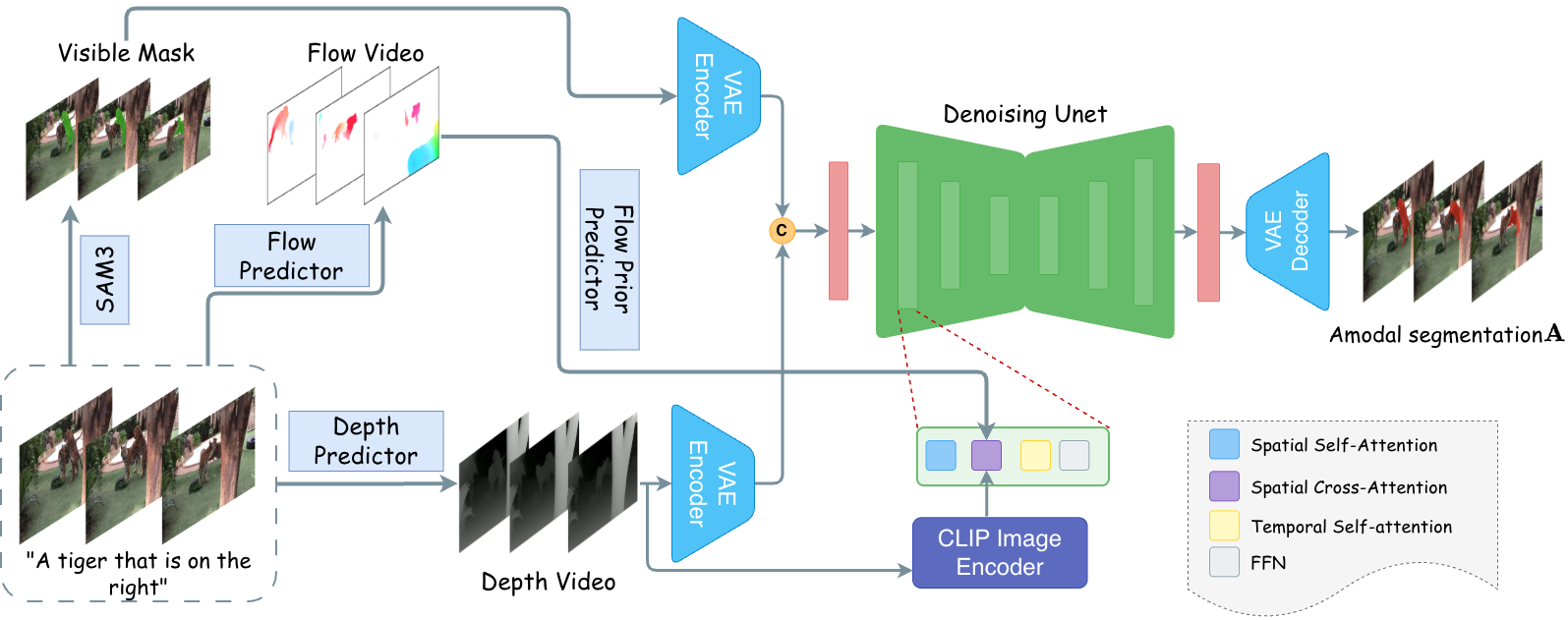}
\caption{\textbf{Video Amodal Mask Segmentation Stage.} The model uses optical flow as a motion prior, warping visible masks from previous frames to propagate shape information into occluded regions.}
    \label{fig:stage1_pipeline}
\end{figure*}
\vspace{-8pt}

\begin{figure*}[!t]
    \centering \includegraphics[width=1.\textwidth]{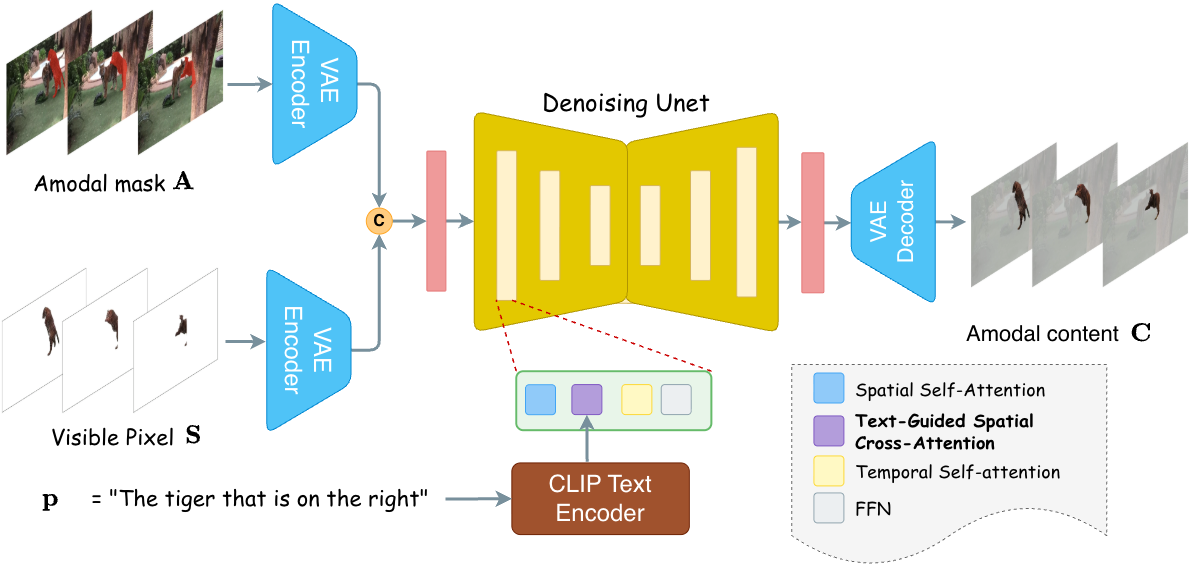}
\caption{\textbf{Video Amodal Texture Completion Stage.} The second stage conducts amodal texture completion with text guidance as semantic priors, leveraging pretrained diffusion models to reconstruct accurate content based on canonical object shapes and properties.}
    \label{fig:stage2_pipeline}
\end{figure*}

\subsection{\textbf{\model}}
Our proposed \model includes two sequential stages: \textit{Video Amodal Mask Segmentation with Optical Flow Guidance}, and \textit{Video Amodal Texture Completion with Text Guidance}. 
The two stages are equipped with prior knowledge to address occlusion. Specifically,
in the first stage, optical flow serves as a motion prior, warping visible masks from previous frames to propagate shape information into occluded regions. In the second stage, text guidance provides a semantic prior, leveraging pretrained diffusion models to supply object-class information. These textual features encode canonical shapes and properties, enabling more accurate reconstruction of occluded content.

\subsubsection{Stage 1: Video Amodal Mask Segmentation with Optical Flow Guidance}
As shown in \cref{fig:stage1_pipeline}, we first train the \textit{Video Amodal Mask Segmentation with Optical Flow Guidance} module to predict the video amodal segmentation masks $\mathbf{A}$. 
Given an RGB video $\mathbf{I}$ and its corresponding visible mask $\mathbf{V}$ for the object of interest, 
this module aims to predict the amodal segmentation masks $\mathbf{A}$ that capture both visible and occluded regions of the object.  Our approach builds upon a video diffusion model (e.g., SVD~\citep{blattmann2023stable}) to leverage shape and content priors learned from large-scale pretraining. 
Specifically, the visible masks $\mathbf{V}$ of shape $\mathbb{R}^{N \times H \times W}$ are first passed through a variational autoencoder (VAE)~\citep{kingma2013auto}. 
Since the VAE requires three input channels, we replicate the binary mask across channels, following the single-channel VAE adaptation of \citet{ke2024repurposing,chen2025using}. 
Encoding through the VAE produces a latent tensor of shape $\mathbb{R}^{N \times C_1 \times \frac{H}{s} \times \frac{W}{s}}$, where $C_1$ is the latent channel dimension and $s$ is the VAE spatial downsampling factor. 
This latent, concatenated with a noise tensor of the same shape and the pseudo-depth latent described below, forms the input to our spatio-temporal 3D U-Net~\citep{blattmann2023align}, 
with a final shape of $\mathbb{R}^{N \times 3C_1 \times \frac{H}{s} \times \frac{W}{s}}$.
Additionally, we incorporate CLIP embeddings~\citep{radford2021learning} of the visible masks $\mathbf{V}$ and inject them into the transformer layers via cross-attention. 
This provides temporal information about the visibility of objects across surrounding frames.  
Inspired by \citet{chen2025using}, we further introduce pseudo-depth maps as an additional input in place of raw RGB frames. 
Occlusions are often caused by objects closer to the camera, and pseudo-depth offers stronger implicit cues about potential occluders than RGB images, 
making it a more effective indicator for identifying regions to complete~\citep{chen2025using}. 
To obtain these pseudo-depth maps, we employ the Depth Anything V2 monocular depth estimator~\citep{yang2024depth}, 
which converts RGB frames into pseudo-depth maps $\mathbf{D} \in \mathbb{R}^{N \times 3 \times H \times W}$ that are concatenated with the visible mask latents as extra channels in the diffusion model.  

Finally, we integrate motion cues from visible regions by incorporating optical flow vectors.
By warping a previous frame's visible mask to the current frame, optical flow can propagate visible content into currently occluded areas, providing a direct shape prior for amodal mask prediction.
To extract optical flow, we adopt the state-of-the-art DPFlow model~\citep{morimitsu2025dpflow}, 
which estimates $\mathbf{F} = \{\mathbf{F}_{i \to i+1}\}_{i=0}^{N-2}$, where $\mathbf{F}_{i \to i+1}$ is the flow from frame $\mathbf{I}_i$ to frame $\mathbf{I}_{i+1}$.

Thus, for each frame, we produce a flow prior feature as guidance for the amodal segmentation mask module. For a specific frame with visible mask $\mathbf{V}_k$, we compute $\tilde{\mathbf{V}}_{k-1 \to k}$, which is the warped mask from the previous frame to the current time step. Because the warped mask can recover regions that are currently occluded but were previously visible, it provides a strong motion prior for amodal mask prediction. Moreover, when occlusion and non-rigid deformation occur simultaneously, the flow-warped mask captures the object's shape transformation, offering a reliable cue where static shape priors from earlier frames would fail. \reviseb{In other words, this flow prior targets \emph{dynamic} occlusion, an object occluded while moving or deforming.}
We then design a simple Flow Prior encoder $g$ that encodes the concatenation of the two masks (each replicated to three channels, as for the VAE input) to $\tilde{\mathbf{F}}_{\text{prior}}\in \mathbb{R}^{N\times d}$, where $d$ is the feature dimension. \revise{In practice, $g$ is a lightweight CNN ($\approx640$K parameters): three Conv$\,\text{-}\,$SiLU stride-$2$ blocks ($6\!\to\!64\!\to\!128\!\to\!256$ channels) followed by global average pooling and a linear projection to $d$, producing one token per frame.} The process can be formulated as follows:

\begin{equation}
\begin{split}
\tilde{\mathbf{V}}_{k-1 \to k} &= \operatorname{warp}(\mathbf{V}_{k-1}, \mathbf{F}_{k-1 \to k}),\\
\tilde{\mathbf{F}}_{\text{prior}} &= g\bigl([\tilde{\mathbf{V}}_{k-1 \to k}, \mathbf{V}_k]\bigr)
\end{split}
\label{eq:flow_prior}
\end{equation}

We then inject $\tilde{\mathbf{F}}_{\text{prior}}$ into the transformer through cross-attention, allowing the model to leverage the flow prior context.
Formally, the objective of the denoising model in this stage is:
\begin{equation}
    \min_{\theta_m} \mathbb{E} \big\| \epsilon - \epsilon_{\theta_m}(\mathbf{z}_t\oplus \mathcal{E}(\mathbf{V})\oplus \mathcal{E}(\mathbf{D}) , \mathcal{C}(\mathbf{V}), \tilde{\mathbf{F}}_{\text{prior}}, t) \big\|_2^2
\label{eq:stage1_objective}
\end{equation}
where $\epsilon$ is the sampled Gaussian noise, $\theta_m$ denotes the learnable parameters of the noise prediction network $\epsilon_{\theta_m}$ for video amodal segmentation, $\mathbf{z}_t$ is the noisy latent at diffusion timestep $t$, $\mathcal{E}(\cdot)$ represents the VAE encoder for visible mask $\mathbf{V}$ and pseudo-depth $\mathbf{D}$, $\mathcal{C}(\cdot)$ denotes the CLIP embedding encoder for visible masks, $\tilde{\mathbf{F}}_{\text{prior}}$ denotes the flow prior feature, $\oplus$ indicates channel-wise concatenation, and $\|\cdot\|_2^2$ is the squared $\ell_2$ norm. At inference, the trained $\theta_m$ is used in DDIM sampling~\citep{song2020denoising} to obtain the predicted video amodal masks $\mathbf{A}$.

\subsubsection{Stage 2: Video Amodal Texture Completion with Text Guidance}
This stage is illustrated in \cref{fig:stage2_pipeline}. After obtaining the video amodal masks $\mathbf{A}$ in the previous module, we input the amodal masks alongside the visible texture of the object of interest $\mathbf{S} = \mathbf{I} \odot \mathbf{V}$. 
Besides these two inputs, we propose a text guidance mechanism for this module. 
The text guidance acts as a semantic prior for completing the occluded texture. 
Utilizing the rich textual representations from pretrained diffusion models, text guidance supplies object-class information that informs the appearance of occluded regions. 
These textual features encode canonical object properties, enabling more accurate reconstruction of hidden content through semantic understanding of the object category.  
In detail, for the conditional latents, we encode both the amodal masks and visible pixels through the VAE to obtain latent vectors. 
These latents, concatenated with a noise latent, form the input of shape $\mathbb{R}^{N \times 3C_1 \times \frac{H}{s} \times \frac{W}{s}}$. 
Additionally, we incorporate CLIP embeddings of the visible pixels $\mathbf{S}$ for cross-attention, which introduces frame-specific appearance cues.  

For the text guidance, we use the CLIP encoder to process the text prompt $\mathbf{p}$ and obtain text embeddings. 
These embeddings are injected into the transformer via cross-attention, propagating semantic priors across frames for content completion.
Because the text prompt remains constant across all frames, it provides a stable semantic reference that helps maintain visual coherence throughout the reconstruction.
By combining visual cues from $\mathbf{S}$ and semantic priors from $\mathbf{p}$, the model gains both appearance-specific and class-level knowledge, leading to more precise and consistent reconstructions.
Formally, the objective for this stage is:
\begin{equation}
    \min_{\theta_c} \mathbb{E} \big\| \epsilon - \epsilon_{\theta_c}(
    \mathbf{z}_t \oplus \mathcal{E}(\mathbf{A}) \oplus \mathcal{E}(\mathbf{S}), 
    \mathcal{C}(\mathbf{S}), 
    \mathcal{C}(\mathbf{p}), 
    t) \big\|_2^2
\end{equation}

where $\theta_c$ are the parameters of the denoising model for content completion, 
$\mathbf{z}_t$ is the noisy latent at timestep $t$, 
$\mathcal{E}(\cdot)$ denotes the VAE encoder applied to the amodal mask $\mathbf{A}$ and visible texture $\mathbf{S}$, 
$\oplus$ indicates channel-wise concatenation, 
$\mathcal{C}(\mathbf{S})$ is the CLIP embedding of the visible pixels, 
$\mathcal{C}(\mathbf{p})$ is the CLIP embedding of the text prompt, 
and $\|\cdot\|_2^2$ is the squared $\ell_2$ distance between the predicted and true noise.
In inference, the trained $\theta_c$ is used in DDIM sampling~\citep{song2020denoising} to obtain the video amodal content $\mathbf{C}$.


\noindent
\textbf{Long Video Inference. } Although current video generative models~\citep{blattmann2023stable, yang2024cogvideox, wan2025wan, chen2025using, lu2025taco} and our \model can theoretically process videos of arbitrary length, significant degradation in completion quality occurs when generating videos whose total length exceeds $N$~\citep{zhang2024avid}. We thus apply Temporal Generation, the long video inference mechanism of \citet{zhang2024avid}, where a sliding window of size $N$ is introduced that incrementally synthesizes future frames based on previously generated outputs.

\subsection{\dataset Benchmark}
\label{subsec:syn_dataset}
We introduce \dataset, a benchmark for text-guided video amodal completion that features long video sequences paired with text descriptions. \dataset includes two subsets: (i) \dataset-S, a large-scale synthetic-occlusion benchmark constructed by compositing objects to create controlled occlusions with amodal ground truth; and (ii) \dataset-R, a small real-occlusion subset of 100 videos, where each video is provided with a human-annotated visible segmentation mask of the object of interest and a corresponding human-written text hint. Existing benchmarks largely overlook long-horizon evaluation and typically contain only short clips, e.g., 24 frames in Kubric~\citep{greff2022kubric} and 14--77 frames in TACO-OvO~\citep{lu2025taco}. Moreover, none of them provide text descriptions; since amodal completion is a generative task, text descriptions enable complementary evaluation via text--video alignment metrics in addition to pixel-level similarity.

\begin{table}[!t]
\centering
\caption{Comparison of statistics with the existing video amodal completion datasets. For \dataset, we report values for the synthetic subset (S) and real subset (R). \# denotes ``number of''.}
\begin{tabular}{l|cccc}
    \toprule
        ~ & \shortstack{Kubric} & \shortstack{TACO-OvO} & \shortstack{\dataset(S)} & \shortstack{\dataset(R)} \\ \midrule
        Video length & Short & Short & Long & Long \\
        \text{  } $\cdot$ Min \#frames & 24 & 14 & 91 & 41 \\ 
        \text{  } $\cdot$ Max \#frames & 24 & 77 & 1595 & 661 \\ 
        \text{  } $\cdot$ Mean \#frames & 24 & 29 & 257 & 151 \\ 
        \# videos & 690 & 2049 & 2,658 & 100 \\ 
        \# total frames & 16,560 & 58,695 & 684,037 & 15,139 \\ 
        Occlusion rate (\%) & 22.96 & 46.67 & 40.21 & - \\
        Text description & \xmark & \xmark & \cmark & \cmark \\ \bottomrule
    \end{tabular}
    \label{tab:dataset_stats}
\end{table}
To construct \dataset-S, we source videos from multiple video object segmentation datasets, including DAVIS~\citep{pont20172017}, YouTube-VOS~\citep{xu2018youtube}, and LVOS~\citep{hong2023lvos}. These sources provide medium-to-long sequences with diverse object types and motion patterns, and often contain a single target object, which simplifies selecting unoccluded object instances to serve as amodal ground truth. We isolate object pixels using the provided masks and synthesize occluded inputs by compositing objects to create controlled occlusion scenarios, while retaining the original unoccluded object pixels as ground-truth completion targets.
For text descriptions, we caption a representative frame (the first frame of the ground-truth video) using a Qwen-VL model (Qwen2.5/Qwen3)~\citep{bai2025qwen2} with a structured instruction that enforces a compact description of the target object. The model is prompted to output a JSON object containing (i) a short noun phrase for the object category, (ii) visible attributes (e.g., color, material, shape), and (iii) the action if applicable. We then convert this structured output into a concise sentence (object + attributes + action) and perform manual refinement by human annotators for correctness.
\cref{tab:dataset_stats} summarizes \dataset and compares it with Kubric~\citep{greff2022kubric} and TACO-OvO~\citep{lu2025taco}. Compared with Kubric and TACO, \dataset exhibits a wider range of video lengths with substantial long video sequences. While Kubric and TACO contain only short clips with maximum lengths of 24 and 77 frames respectively, \dataset supports long sequences, with up to 1,595 frames in the synthetic subset and 661 frames in the real subset. The average sequence length in \dataset, 257 frames for synthetic and 151 frames for real, is significantly higher than Kubric, 24 frames, and TACO, 29 frames, demonstrating its suitability for long-term temporal reasoning.
In terms of scale, \dataset is also considerably larger. The synthetic subset contains 2,658 videos and 684,037 frames, exceeding Kubric, 690 videos and 16,560 frames, and TACO, 2,049 videos and 58,695 frames, by a large margin. The real subset further contributes 100 videos and 15,139 frames, providing additional real-world diversity. Moreover, \dataset exhibits a high occlusion rate of 40.21 percent, making it more challenging and realistic for amodal completion tasks. Importantly, unlike prior datasets, \dataset provides textual descriptions for both synthetic and real subsets, enabling multimodal learning and text-guided amodal completion, which is not supported in Kubric or TACO.

\section{Experimental Results}
\subsection{Experimental Setup}

\begin{table*}[!t]
\small
\centering
\setlength{\tabcolsep}{3pt}
\caption{{Quantitative comparison with state-of-the-art methods.} Bold indicates the best performance.}
\resizebox{\linewidth}{!}{
\begin{tabular}{c|l|cccc|c|cc}
\toprule
\multirow{2}[2]{*}{\textbf{Dataset}} & \multirow{2}[2]{*}{\textbf{Method}} & \multicolumn{4}{c|}{\textbf{Texture Completion}} & \multicolumn{1}{c|}{\textbf{Segmentation}} & \multicolumn{2}{c}{\textbf{Temporal Consistency}} \\
\cmidrule(lr){3-6} \cmidrule(lr){7-7} \cmidrule(lr){8-9}
 &  & PSNR($\uparrow$) & SSIM($\uparrow$) & LPIPS($\downarrow$) & CLIP-T($\uparrow$)  & mIoU($\uparrow$) & CLIP-F($\uparrow$) & FVD($\downarrow$)\\
\midrule
\multirow{8}{*}{\shortstack{Kubric}} &
          Pix2Video & 14.876 & 0.704 & 0.325 & - & 27.45 & 0.939 & 1073.13 \\ 
        ~ & E2FGVI & 14.406 & 0.688 & 0.227 & - & 54.05 & 0.967 & 732.30 \\ 
        ~ & pix2gestalt & 20.096 & 0.826 & 0.129 & -  & 71.95 & 0.911 & 405.89 \\ 
        ~ & PCNet & 20.111 & 0.861 & 0.111 & -  & 79.70 & 0.958 & 253.77 \\ 
        ~ & Diff-VAS & 21.213 & 0.852 & 0.099 & -  & 81.05 & 0.974 & 229.04 \\
        ~ & TACO & 23.509 & 0.889 & 0.077 & -  & 80.65 & 0.970 & 186.10 \\ 
        ~ & \textbf{\model (Ours)} & \textbf{29.916} & \textbf{0.921} & \textbf{0.051}& -  & \textbf{84.12} & \textbf{0.975} & \textbf{122.66} \\
\midrule
\multirow{7}{*}{\shortstack{TACO-\\OvO}}
          & Pix2Video & 12.231 & 0.672 & 0.385 & - & 38.65 & 0.941 & 1638.49 \\
        ~ & E2FGVI & 13.123 & 0.652 & 0.326 & - & 59.65 & 0.966 & 1137.89 \\ 
        ~ & pix2gestalt & 19.442 & 0.803 & 0.133 & - & 86.30 & 0.920 & 532.01 \\ 
        ~ & TACO & 24.342 & 0.882 & 0.074 & - & 94.00 & 0.978 & 216.93 \\ 
        ~ & \textbf{\model (Ours)} & \textbf{26.881} & \textbf{0.899} & \textbf{0.068} & - & \textbf{95.12} &\textbf{0.988} & \textbf{213.77} \\
        \midrule
\multirow{7}{*}{\shortstack{\dataset-S\\(Ours)}}
        ~ & pix2gestalt  & 21.057 & 0.952 & 0.172 &  0.279 & 83.83 & 0.914 & 555.95\\ 
        ~ & VACE & 23.541 & 0.943 & 0.106 & 0.270 & 59.09 & 0.892 & 960.14\\ 
        ~ & Diff-VAS & 23.933 & 0.948 & 0.081 & 0.271 & 86.60 & 0.935 & 443.59 \\ 
        ~ & TACO & 24.821 & 0.951 & 0.082 & 0.272 & 85.28 & 0.925 & 471.42  \\ 
        ~ & \textbf{\model (Ours)} & \textbf{27.473} & \textbf{0.963} & \textbf{0.073} & \textbf{0.285} & \textbf{90.29} & \textbf{0.945} & \textbf{310.90} \\  
\bottomrule
\end{tabular}
}
\label{tab:quant_compare_kubric_taco_lavat}
\end{table*}

\begin{table}[!t]
\centering
\setlength{\tabcolsep}{20pt}
\caption{{Evaluation on \dataset-R (real occlusions).} We report text-video alignment (CLIP-T) and human preference (\%) from a user study.}
\begin{tabular}{l|cc}
\toprule
\textbf{Method} & CLIP-T($\uparrow$) & User Pref. ($\uparrow$) \\
\midrule
pix2gestalt~\citep{ozguroglu2024pix2gestalt} & 0.274 & 6\% \\
Diff-VAS~\citep{chen2025using} & 0.279 & 12\% \\
TACO~\citep{lu2025taco} & 0.280 & 11\% \\
\textbf{\model (Ours)} & \textbf{0.298} & \textbf{71\%} \\
\bottomrule
\end{tabular}
\label{tab:lavatR}
\end{table}

\noindent
\textbf{Implementation Details.}
For training, we follow a similar training setting to Diff-VAS~\citep{chen2025using} for fair comparison. We initialize from the official SVD-xt~\citep{blattmann2023stable} pretrained checkpoint and finetune on curated SAILVOS dataset from \citet{chen2025using}. We adopt the AdamW optimizer with a learning rate of $3 \times 10^{-6}$, a batch size of 8, and a frame resolution of $128 \times 256$. Training requires approximately 24 hours on 8 NVIDIA RTX A6000 GPUs. During inference, we use 25 DDIM~\citep{song2020denoising} denoising steps with a guidance scale of 4.5. We apply a sliding window of size 25 with a stride of 4, following \citet{zhang2024avid}.

\noindent
\textbf{Datasets.} We evaluate on three benchmarks: our {\dataset}, Kubric~\citep{greff2022kubric}, and TACO-OvO~\citep{lu2025taco}. To enable text guidance on Kubric and TACO-OvO, we generate text descriptions for the amodal texture ground truth using Qwen2.5~\citep{bai2025qwen2}, followed by manual refinement.

\noindent
\textbf{Metrics.} We evaluate texture completion using PSNR~\citep{gonzalez2009digital}, SSIM~\citep{wang2004image}, and LPIPS~\citep{zhang2018unreasonable} for image-level alignment. For \dataset, we additionally report CLIP-T~\citep{radford2021learning} to assess text--video alignment. Amodal segmentation quality is measured by mIoU. For temporal consistency, we use CLIP-F~\citep{esser2023structure} for frame-to-frame similarity and FVD~\citep{unterthiner2018towards} for sequence-level coherence. For \dataset-R, we report CLIP-T and human preference (\%) from a user study.

\begin{figure}[!t]
    \centering
    \includegraphics[width=1\linewidth]{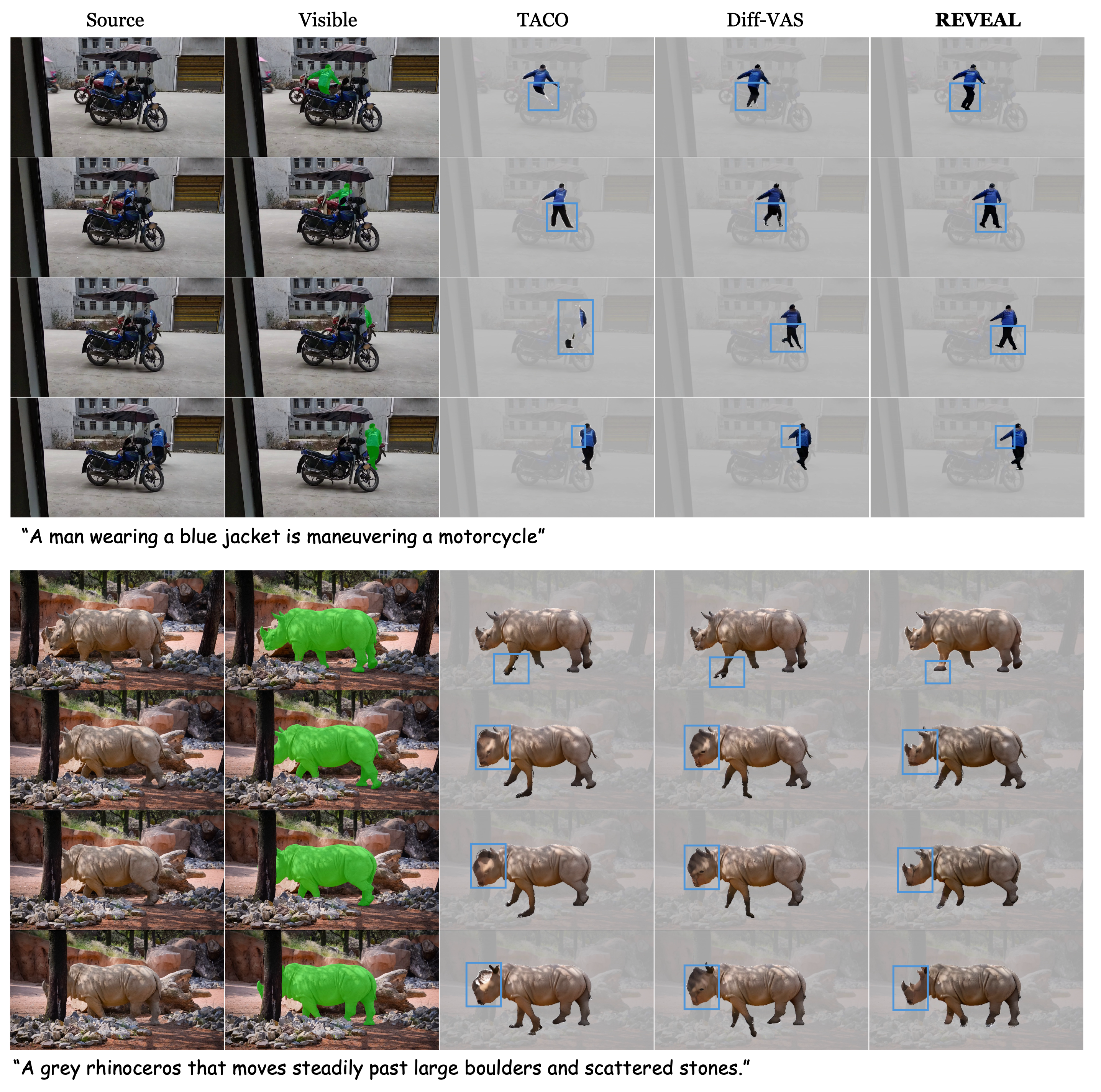}
    \caption{Qualitative comparison between TACO~\citep{lu2025taco}, Diff-VAS~\citep{chen2025using}, and our {\model} on \dataset-R benchmark.}
    \label{fig:quali_itw}
\end{figure}

\subsection{Comparison to State-of-the-art}
We compare against recent video amodal completion methods Diff-VAS~\citep{chen2025using} and TACO~\citep{lu2025taco}, image-level amodal completion models pix2gestalt~\citep{ozguroglu2024pix2gestalt} and PCNet~\citep{zhan2020self} extended frame-by-frame, video inpainting methods E2FGVI~\citep{li2022towards} and VACE~\citep{jiang2025vace}, and a video editing approach Pix2Video~\citep{ceylan2023pix2video}.

\noindent
\textbf{Quantitative Results.} As shown in \cref{tab:quant_compare_kubric_taco_lavat}, \model achieves state-of-the-art performance across all three benchmarks. On Kubric, \model achieves substantial gains in texture completion (PSNR 29.92, SSIM 0.921, LPIPS 0.051) over the strongest baseline TACO (23.51 PSNR, 0.077 LPIPS), while also improving segmentation (mIoU 84.12) and reducing temporal artifacts (FVD 122.66). On TACO-OvO, \model improves both completion and segmentation (PSNR 26.88, LPIPS 0.068, mIoU 95.12) with stronger temporal consistency (CLIP-F 0.988, FVD 213.77). On our long-horizon \dataset-S, \model achieves the best performance across all metrics, including CLIP-T (0.285), mIoU (90.29), and FVD (310.90), demonstrating the effectiveness of combining motion and semantic priors under occlusion.
\dataset-R contains 100 real-occlusion videos with human-annotated visible masks and text hints. Since amodal ground truth is unavailable, we evaluate via CLIP-T and a user study. As shown in \cref{tab:lavatR}, \model achieves the best CLIP-T and highest user preference (71\%), indicating more semantically aligned and perceptually preferred completions.

\noindent
\textbf{Qualitative Results.} \cref{fig:quali_itw} illustrates the qualitative advantage of \model on \dataset-R. In the first example, a man on a motorcycle is heavily occluded; TACO and Diff-VAS produce incomplete body and limb regions, whereas \model reconstructs a more complete and coherent figure. In the second example, a rhinoceros is partially hidden behind boulders; competing methods generate incomplete shapes and artifacts, while \model recovers the full body with consistent texture across frames. \revise{Qualitative comparisons against ground-truth amodal content on Kubric, TACO-OvO, and \dataset-S, along with additional qualitative results, are provided in the supplementary material. Video results are available on the paper webpage.}

\begin{table}[!t]
\small
\centering
\setlength{\tabcolsep}{4pt}
\caption{\textbf{Flow prior ablation across occlusion difficulty on \dataset-S.} We report amodal mIoU ($\uparrow$).}
\begin{tabular}{l|cccc}
\toprule
\textbf{Setting} & $<20\%$ & $20$--$40\%$ & $40$--$65\%$ & $>65\%$ \\
\midrule
w/o flow prior & 91.88 & 87.97 & 85.75 & 87.79 \\
w/ flow prior  & 92.93 & 90.72 & 88.38 & 89.13 \\
\bottomrule
\end{tabular}
\label{tab:abla_flow_occl}
\end{table}

\begin{figure}[!t]
    \centering
    \includegraphics[width=1\linewidth]{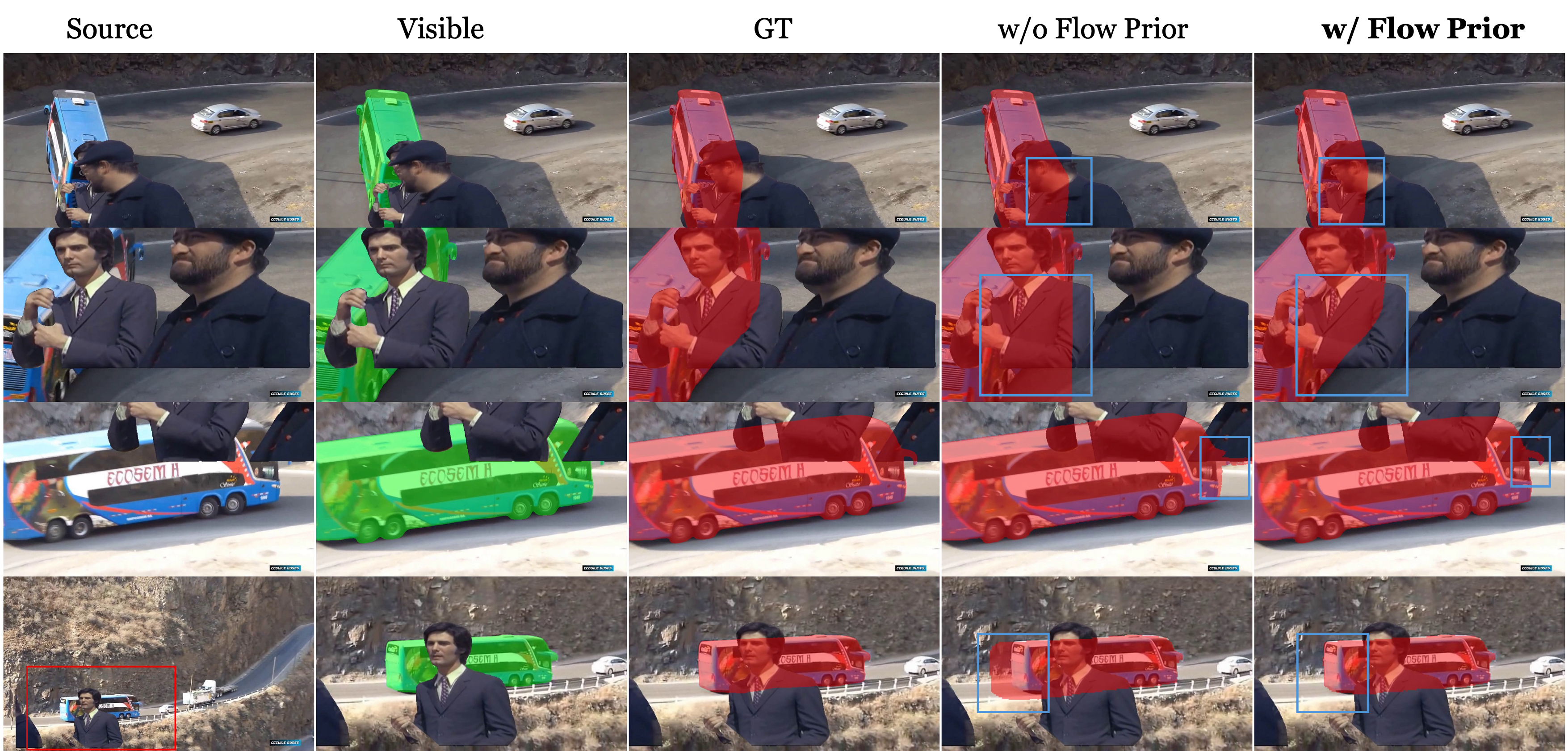}
    \caption{\textbf{Ablation of flow prior.} Amodal mask prediction with and without flow prior on \dataset.}
    \label{fig:quali_abla_flow}
    \vspace{-2mm}
\end{figure}

\begin{figure}[!t]
    \centering
    \includegraphics[width=1\linewidth]{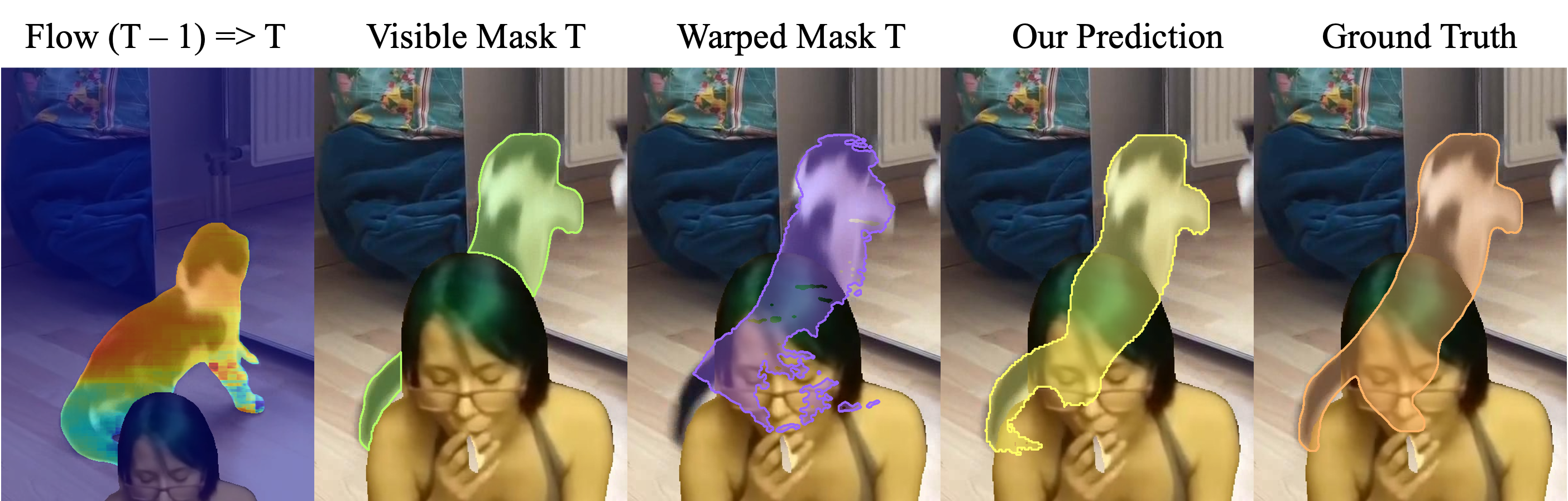}
    \caption{\textbf{Illustration of the flow prior under simultaneous deformation and occlusion.} The object deforms while being occluded. The visible mask at frame T (green) covers only the unoccluded portion. Warping the previous frame's mask to T via optical flow (purple) recovers regions that are currently hidden, providing a shape prior that captures the object's motion and deformation. Our predicted amodal mask (yellow) closely matches the ground truth (orange).}
    \label{fig:flow_prior_details}
\end{figure}

\subsubsection{Ablation Study}
\noindent
\textbf{Effect of flow mask priors guidance.} \cref{tab:abla_flow_occl} shows that the optical-flow mask prior consistently improves amodal segmentation across all occlusion levels, with gains that scale with occlusion severity. Under light occlusion ($<20\%$), the improvement is modest ($+1.05$ mIoU), as the visible mask already covers most of the object and leaves little ambiguity about its spatial extent. However, the benefit increases substantially under heavier occlusion: mIoU improves by $+2.75$ for $20$--$40\%$ occlusion ($87.97 \to 90.72$) and by $+2.63$ for $40$--$65\%$ occlusion ($85.75 \to 88.38$). Under extreme occlusion ($>65\%$), the flow prior still yields a meaningful gain of $+1.34$ mIoU ($87.79 \to 89.13$). This trend aligns with our hypothesis: when more of the object is hidden, visible masks from previous frames---warped via optical flow---provide increasingly valuable evidence about the locations of occluded regions. \cref{fig:quali_abla_flow} confirms that flow priors help recover missing regions with cleaner boundaries, especially when the visible mask is fragmented. \cref{fig:flow_prior_details} illustrates the case where the object simultaneously deforms and becomes occluded. The visible mask at frame T only covers the unoccluded portion, yet warping the previous frame's mask via optical flow recovers the hidden regions by capturing the object's motion and shape change, producing a prior that closely approximates the ground-truth amodal mask.

\begin{table}[!t]
\small
\centering
\setlength{\tabcolsep}{4pt}
\caption{\reviseC{\textbf{Motion guidance ablation across motion and shape-change magnitudes on \dataset-S.} We report amodal mIoU ($\uparrow$), grouping motion and shape change separately.}}
\label{tab:abla_flow_motion}
\reviseC{%
\begin{tabular}{l|ccc|ccc}
\toprule
 & \multicolumn{3}{c|}{Motion} & \multicolumn{3}{c}{Shape change} \\
\cmidrule(lr){2-4} \cmidrule(lr){5-7}
\textbf{Setting} & Low & Medium & High & Low & Medium & High \\
\midrule
w/o motion guidance & 89.55 & 85.28 & 76.26 & 90.36 & 85.72 & 77.70 \\ w/ motion guidance (Ours) & \textbf{92.48} & \textbf{86.84} & \textbf{78.10} & \textbf{93.30} &       
  \textbf{87.87} & \textbf{80.73} \\ 
\bottomrule
\end{tabular}%
}
\end{table}

\reviseC{\noindent
\textbf{Effect of motion and shape-change magnitude.} We evaluate motion and shape change separately on \dataset-S using ground-truth amodal masks. Motion magnitude is $m_t=\|c_t-c_{t-1}\|_2/d_{t-1}$, where $c_t$ is the mask centroid at frame $t$ and $d_{t-1}$ is the previous-frame object's bounding-box diagonal. Shape-change magnitude is $s_t=1-\operatorname{IoU}(\operatorname{align}(A_{t-1}^{\mathrm{GT}}),A_t^{\mathrm{GT}})$, where alignment translates the previous mask to match the current mask's centroid. Low, medium, and high motion correspond to $m_t\leq0.05$, $0.05<m_t\leq0.20$, and $m_t>0.20$; the corresponding shape-change groups are $s_t\leq0.10$, $0.10<s_t\leq0.50$, and $s_t>0.50$. As shown in \cref{tab:abla_flow_motion}, our motion guidance improves mIoU across all motion and shape-change groups. It remains beneficial under high motion and high shape change, with gains of  1.84 and 3.03, respectively.}

\begin{table}[!t]
\small
\centering
\setlength{\tabcolsep}{4pt}
\caption{\revise{\textbf{Flow-prior injection design ablation on \dataset-S.} Added params is on top of the 397.66M-parameter Stage-1 trainable set; Latency is per 25-frame 128$\times$256 fp16 step on a single A6000. mIoU ($\uparrow$) is reported across occlusion difficulty.}}
\label{tab:abla_flow_injection}
\begin{tabular}{l|cc|cccc}
\toprule
\textbf{Flow injection} & \textbf{Added params} & \textbf{Latency (ms)} & $<20\%$ & $20$--$40\%$ & $40$--$65\%$ & $>65\%$ \\
\midrule
w/o flow prior      & \revise{0}        & \revise{86.9}          & \revise{91.88}        & \revise{87.97}        & \revise{85.75}        & \revise{87.79} \\
Latent concat       & \revise{11.5K}    & \revise{87.2 (+0.3)}   & \revise{92.20}        & \revise{88.50}        & \revise{86.40}        & \revise{88.10} \\
MOFA-Adapter        & \revise{735.2M}   & \revise{116.4 (+29.5)} & \revise{92.80}        & \revise{90.30}        & \revise{87.95}        & \revise{88.80} \\
\textbf{Ours}       & \revise{0.635M} & \revise{90.2 (+3.3)} & \revise{\textbf{92.93}} & \revise{\textbf{90.72}} & \revise{\textbf{88.38}} & \revise{\textbf{89.13}} \\
\bottomrule
\end{tabular}
\end{table}

\revise{\noindent\textbf{Flow-prior injection design.} We compare three strategies for injecting the warped-mask flow signal into the Stage-1 U-Net, all using the same measured dense flow $\mathbf{F}_{k-1\to k}$ from DPFlow and identical training settings. (i) \emph{Latent concat}: VAE-encode the warped previous-frame mask $\tilde{\mathbf{V}}_{k-1 \to k}$ and append it as an extra input-channel block, so the input becomes $\mathbf{z}_t \oplus \mathcal{E}(\mathbf{V}) \oplus \mathcal{E}(\mathbf{D}) \oplus \mathcal{E}(\tilde{\mathbf{V}}_{k-1 \to k})$. (ii) \emph{MOFA-Adapter-style}, following MOFA-Video~\citep{niu2024mofa}: a multi-scale Reference Encoder over $\mathbf{V}_{k-1}$, forward-warping under $\mathbf{F}_{k-1\to k}$, a Fusion Encoder mirroring the SVD encoder, and zero-conv residuals added into the frozen SVD decoder at every scale. (iii) \emph{Ours}: a lightweight CNN $g$ encodes the warped previous-frame mask (concatenated with $\mathbf{V}_k$) into a single per-frame token $\tilde{\mathbf{F}}_{\text{prior}}$ (\cref{eq:flow_prior}), injected into the Stage-1 U-Net as a cross-attention KV stream (\cref{eq:stage1_objective}). \cref{tab:abla_flow_injection} reports amodal mIoU on \dataset-S by occlusion difficulty, with added trainable parameters and per-step latency. Cross-attention obtains the highest mIoU in every occlusion bucket at $0.635$M added parameters and $+3.3$ ms latency. Latent concat yields only marginal gains over the no-flow baseline despite the lowest cost, and MOFA-Adapter-style improves over latent concat but remains below cross-attention in every bucket while adding $735.2$M parameters and $+29.5$ ms latency.}

\begin{table}[!t]
\small
\centering
\setlength{\tabcolsep}{4pt}
\caption{\textbf{Text guidance ablation across occlusion difficulty on \dataset-S.} We report completion quality for our Stage-2 texture model with and without text guidance.}
\resizebox{1.0\textwidth}{!}{
\begin{tabular}{l|ccc|ccc|ccc|ccc}
\toprule
 & \multicolumn{3}{c|}{$<20\%$} & \multicolumn{3}{c|}{$20$--$40\%$} & \multicolumn{3}{c|}{$40$--$65\%$} & \multicolumn{3}{c}{$>65\%$} \\
\cmidrule(lr){2-4} \cmidrule(lr){5-7} \cmidrule(lr){8-10} \cmidrule(lr){11-13}
\textbf{Setting} &
SSIM$\uparrow$ & LPIPS$\downarrow$ & CLIP-T$\uparrow$ &
SSIM$\uparrow$ & LPIPS$\downarrow$ & CLIP-T$\uparrow$ &
SSIM$\uparrow$ & LPIPS$\downarrow$ & CLIP-T$\uparrow$ &
SSIM$\uparrow$ & LPIPS$\downarrow$ & CLIP-T$\uparrow$\\
\midrule
w/o text guidance  & 0.982 & 0.026 & 0.281 & 0.945 & 0.087 & 0.282 & 0.945 & 0.087 & 0.279 & 0.942 & 0.090 & 0.269 \\
w/ text guidance   & 0.987 & 0.021 & 0.286 & 0.960 & 0.067 & 0.285 & 0.959 & 0.071 & 0.284 & 0.957 & 0.072 & 0.273 \\
\bottomrule
\end{tabular}
}
\label{tab:abla_text_occl}
\vspace{-3mm}
\end{table}

\begin{figure}[!t]
    \centering
    \includegraphics[width=1\linewidth]{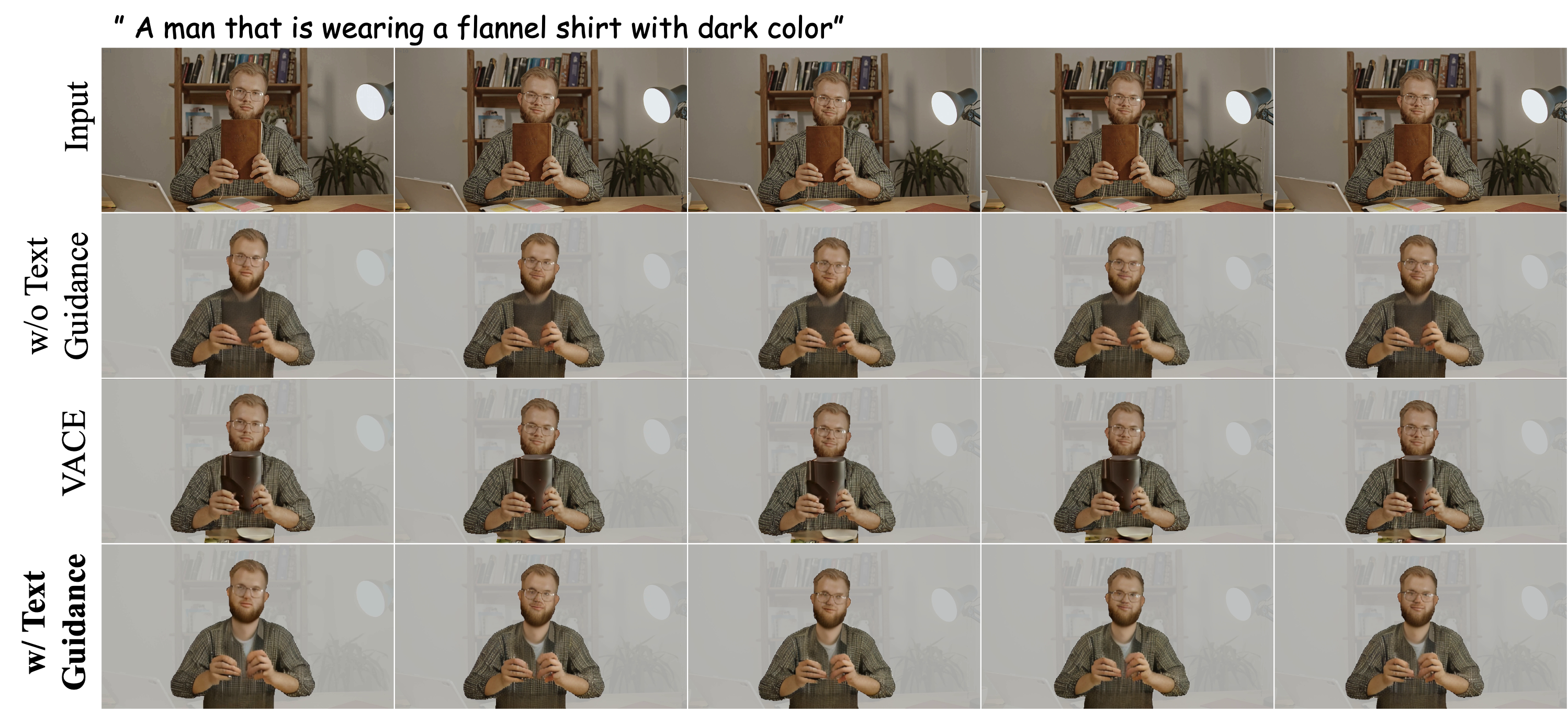}
    \caption{\textbf{Qualitative ablation of text guidance.} Texture completion with and without text guidance on \dataset. We also compare with VACE~\citep{jiang2025vace} as an additional video inpainting baseline.}
    \label{fig:quali_abla_text}
    \vspace{-4mm}
\end{figure}

\begin{figure}[!ht]
    \centering
    \includegraphics[width=1\linewidth]{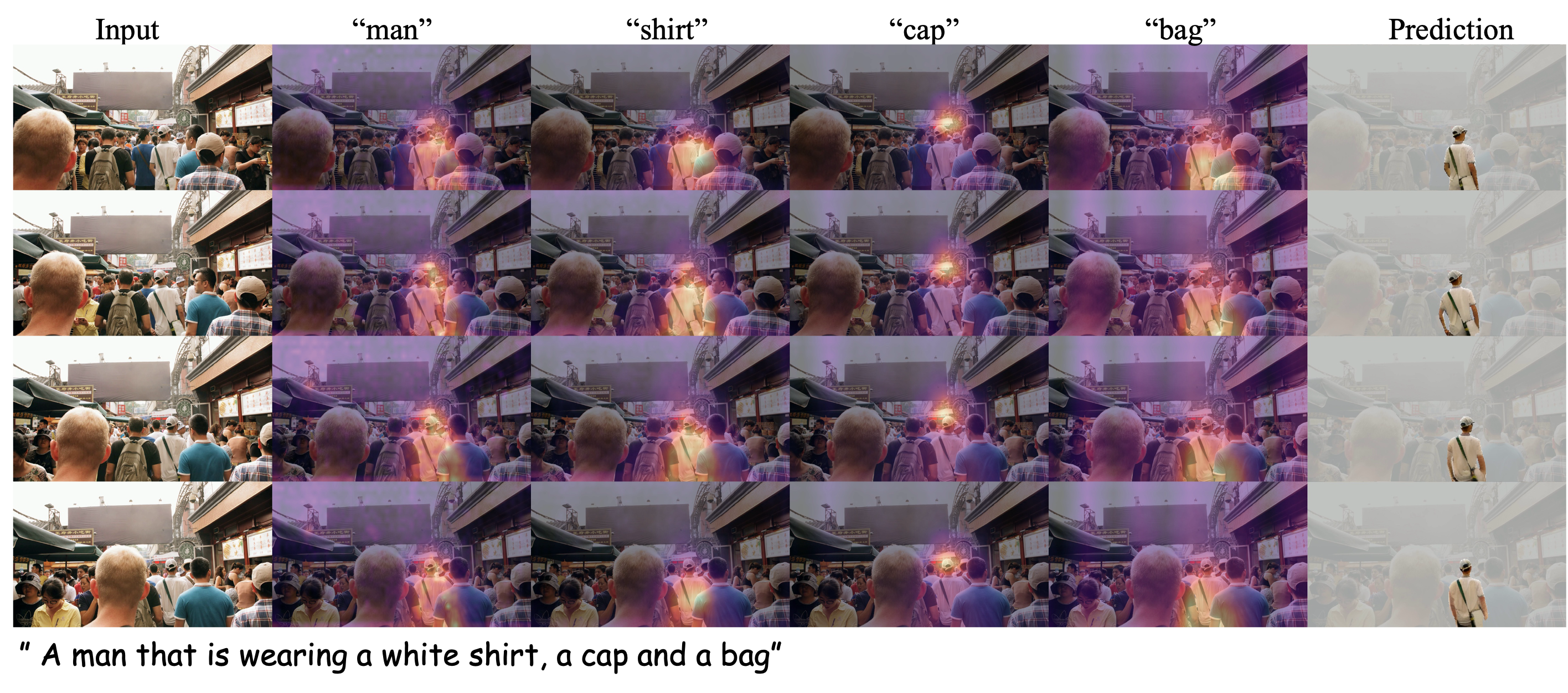}
    \caption{\textbf{Text cross-attention visualization.} Given the prompt ``A man that is wearing a white shirt, a cap and a bag'', we visualize the cross-attention maps for individual tokens. Each token attends to its corresponding semantic region, demonstrating that text guidance provides spatially grounded priors.}
    \label{fig:text_attn_viz}
    \vspace{-3mm}
\end{figure}

\noindent
\textbf{Effect of textual prior guidance.}
\cref{tab:abla_text_occl} demonstrates that text guidance improves completion quality across all occlusion regimes, with the benefit increasing as occlusion grows more severe. Under light occlusion ($<20\%$), the model already reconstructs well from visible content alone, and text guidance provides only marginal gains (SSIM $+0.005$, LPIPS $-0.005$). As occlusion increases, the improvement grows substantially: at $20$--$40\%$ occlusion, SSIM improves by $+0.015$ and LPIPS decreases by $0.020$; at $40$--$65\%$, SSIM improves by $+0.014$ and LPIPS by $0.016$. Under heavy occlusion ($>65\%$), text guidance improves SSIM from $0.942$ to $0.957$ and reduces LPIPS from $0.090$ to $0.072$, yielding the largest absolute gains. This trend validates our core motivation: when visible content is insufficient to infer the complete appearance, text priors supply the missing semantic knowledge about the object's canonical shape and properties, leading to more faithful reconstructions. Qualitatively, \cref{fig:quali_abla_text} shows that the unguided variant produces incomplete or washed-out regions. VACE~\citep{jiang2025vace}, as a video inpainting method, tends to fill occluded regions based on surrounding context rather than reconstructing the object as a whole, resulting in completions that blend with the background instead of preserving the object's own appearance. In contrast, our model with text guidance leverages semantic priors about the object category (e.g., ``a man wearing a flannel shirt with dark color''), enabling coherent object reconstruction across frames. \cref{fig:text_attn_viz} further visualizes the cross-attention maps for individual text tokens, showing that each token (e.g., ``man'', ``shirt'', ``cap'') attends to the corresponding semantic region, confirming that text guidance provides spatially grounded priors.

\begin{table}[!t]
\small
\centering
\setlength{\tabcolsep}{4pt}
\caption{\reviseC{\textbf{Temporal consistency of text guidance across sequence lengths on \dataset-S.} We report Stage-2 texture completion results with and without text guidance using CLIP-F ($\uparrow$) and FVD ($\downarrow$). The with-text results are also reported in the ``w/ Temporal Gen.'' row of \cref{tab:abla_temporal_length}.}}
\label{tab:abla_text_length}
\reviseC{%
\begin{tabular}{l|cc|cc|cc}
\toprule
 & \multicolumn{2}{c|}{$<250$ frames} & \multicolumn{2}{c|}{250--500 frames} & \multicolumn{2}{c}{$>500$ frames} \\
\cmidrule(lr){2-3} \cmidrule(lr){4-5} \cmidrule(lr){6-7}
\textbf{Text guidance} & CLIP-F$\uparrow$ & FVD$\downarrow$ & CLIP-F$\uparrow$ & FVD$\downarrow$ & CLIP-F$\uparrow$ & FVD$\downarrow$ \\
\midrule
w/o text guidance &  0.942 & 244.50 &
 0.943 & 348.80 &
 0.928 & 392.70 \\
w/ text guidance &  0.952 & 225.10 & 0.945 &303.12 &  0.940 & 342.55 \\
\bottomrule
\end{tabular}%
}
\end{table}

\reviseC{\noindent
\textbf{Temporal consistency of text guidance.} \cref{tab:abla_text_length} evaluates Stage-2 texture completion with and without text guidance on \dataset-S across three sequence-length groups: fewer than $250$ frames, $250$--$500$ frames, and more than $500$ frames. We report CLIP-F for frame-to-frame similarity and FVD for sequence-level video consistency. Text guidance improves both metrics in every group: from the shortest to the longest group, CLIP-F increases by 0.010, 0.002, and 0.012, while FVD decreases by 19.40, 45.68, and 50.15, respectively. The largest absolute gains occur for videos longer than 500 frames, where CLIP-F improves from 0.928 to 0.940 and FVD decreases from 392.70 to 342.55. This finding is consistent with our motivation that a fixed text prompt provides a stable semantic reference across frames.}

\begin{table}[!t]
\small
\centering
\setlength{\tabcolsep}{4pt}
\caption{\reviseb{\textbf{Flow prior ablation across occlusion duration on \dataset-S.} Amodal mIoU ($\uparrow$) over occluded frames, grouped by the number of consecutive frames the object has been occluded.}}
\label{tab:abla_flow_duration}
\begin{tabular}{l|cc|c}
\toprule
\textbf{Occlusion duration (frames)} & \textbf{w/o flow prior} & \textbf{w/ flow prior} & $\Delta$ \\
\midrule
1--15  & 87.39 & 89.89 & +2.50 \\
16--40 & 88.03 & 90.61 & +2.58 \\
41--80 & 89.73 & 92.29 & +2.56 \\
81+    & 90.77 & 93.07 & +2.30 \\
\bottomrule
\end{tabular}
\end{table}

\begin{table}[!t]
\small
\centering
\setlength{\tabcolsep}{4pt}
\caption{\reviseb{\textbf{End-to-end evaluation from plain-text input on \dataset-S.} The first row upper-bounds Stage 2 with ground-truth amodal masks; the second row is our reported setting; the third row runs the full pipeline from the text prompt alone, with visible masks predicted by SAM3.}}
\label{tab:end_to_end}
\begin{tabular}{ll|ccccc}
\toprule
\textbf{Visible mask} & \textbf{Amodal mask $\to$ Stage 2} & mIoU$\uparrow$ & PSNR$\uparrow$ & SSIM$\uparrow$ & LPIPS$\downarrow$ & FVD$\downarrow$ \\
\midrule
GT & GT amodal (Stage-2 upper bound) & --- & 29.55 & 0.972 & 0.058 & 276.4 \\
GT & Stage-1 predicted (reported setting) & 90.29 & 27.47 & 0.963 & 0.073 & 310.90 \\
SAM3~\citep{carion2025sam} & Stage-1 predicted (end-to-end from text) & 88.1 & 26.4 & 0.955 & 0.084 & 338.0 \\
\bottomrule
\end{tabular}
\end{table}

\reviseb{\noindent
\textbf{Effect of occlusion duration.} \cref{tab:abla_flow_duration} groups Stage-1 accuracy by the number of consecutive frames an object has been occluded. The flow prior yields a consistent gain of $+2.3$ to $+2.6$ mIoU across all buckets, from transient occlusions (1--15 frames) to long-lasting ones (81+ frames), showing that its benefit is invariant to occlusion duration.}

\reviseb{\noindent
\textbf{End-to-end evaluation from plain-text input.} \cref{tab:end_to_end} quantifies the error accumulation of the full pipeline on \dataset-S. Feeding Stage 2 the Stage-1 predicted amodal mask instead of the ground-truth amodal mask (first vs.\ second row) costs $\approx$2.1 PSNR and $+0.015$ LPIPS. Replacing ground-truth visible masks with SAM3~\citep{carion2025sam} predictions obtained from the text prompt alone (second vs.\ third row), i.e., fully end-to-end from text, incurs only a modest drop of $\approx$1.1 PSNR and $2.2$ mIoU. }

\section{Conclusion \& Discussion}

We presented \model, a diffusion-based framework for open-vocabulary video amodal completion that integrates complementary prior knowledge to address the limitations of existing methods. A user-provided text query serves a dual role: it drives open-vocabulary segmentation to obtain visible masks, and it acts as a semantic prior for texture reconstruction. In the first stage, optical flow serves as a motion prior, warping visible masks from previous frames to propagate shape information into occluded regions, providing a strong shape prior particularly under heavy occlusion and when deformation and occlusion occur simultaneously. In the second stage, text guidance supplies object-class information through cross-attention for more faithful texture reconstruction, while providing a stable semantic reference that helps maintain visual coherence throughout the reconstruction. We also introduced \dataset, a benchmark for text-guided video amodal completion featuring long video sequences paired with text descriptions, enabling evaluation of both temporal consistency and text-video alignment beyond the short clips used in prior work. Extensive experiments on Kubric, TACO-OvO, and \dataset demonstrate that \model achieves state-of-the-art performance across completion fidelity, segmentation accuracy, and temporal consistency.

\revise{\noindent
\textbf{Generalization to DiT-based video diffusion.} Our two priors are architecture-agnostic conditioning signals. The latent-level conditioning $\mathbf{z}_t \oplus \mathcal{E}(\mathbf{V}) \oplus \mathcal{E}(\mathbf{D})$ ports to a video Diffusion Transformer (DiT) such as CogVideoX~\citep{yang2024cogvideox} or WAN-Video~\citep{wan2025wan} by widening the patch embedder's input projection. The flow prior $\tilde{\mathbf{F}}_{\text{prior}}$ maps onto a DiT cross-attention layer as an additional KV stream, and the Stage-2 text prior is already a CLIP-conditioned cross-attention native to DiTs. A DiT instantiation of \model is left to future work.}

\noindent
\textbf{Limitation.} While text guidance improves completion quality by providing language priors, its effectiveness is primarily limited to category-level semantics. The text prompt can reliably guide the reconstruction of general object appearance and, to some extent, fine-grained properties such as texture and color. However, it does not yet consistently resolve ambiguities in more dynamic attributes, such as object pose or action (e.g., distinguishing between sitting and running). \revise{We provide a qualitative illustration of this failure mode in the supplementary material.} \reviseb{In addition, the flow prior requires the object to be visible in preceding frames: its gain is stable across occlusion durations (\cref{tab:abla_flow_duration}), but objects that remain heavily occluded from their first appearance provide no visible region to propagate, leaving such persistent occlusion outside the scope of motion-based priors.}
\reviseC{Moreover, our text guidance uses one fixed object description for the entire sequence, and the descriptions in \dataset-S are derived from the first ground-truth frame. This assumes that the description remains applicable throughout the video; the current model, training strategy, and dataset construction do not explicitly support substantial changes in object state, appearance, or action, such as walking to running or removing a hat. Handling such changes requires identifying when the description becomes inaccurate and which text tokens should guide each part of the sequence, making time-varying text guidance a direction for future work.}

\clearpage
\bibliography{tmlr}
\bibliographystyle{tmlr}

\appendix

\end{document}